\documentclass{article}
\newcommand{\final}{0}

\usepackage{hyperref}
\usepackage{url}
\usepackage{booktabs}
\usepackage{amsfonts}
\usepackage{xcolor}
\usepackage{nicefrac}
\usepackage{microtype}
\usepackage{bm}
\usepackage{amsmath}
\usepackage{amssymb}
\usepackage{soul}
\usepackage{graphicx}
\usepackage{caption}
\usepackage{subcaption}

\definecolor{WeimingColor}{rgb}{0,0,0.8} 
\newcommand{\weiming}[1]{{\color{WeimingColor} [Weiming: #1]}}
\definecolor{YipingColor}{rgb}{0.8,0,0} 
\newcommand{\yiping}[1]{{\color{YipingColor} [Yiping: #1]}}
\definecolor{HuaiyuColor}{rgb}{0.0,0.8,0} 
\newcommand{\huaiyu}[1]{{\color{HuaiyuColor}[Huaiyu: #1]}}
\definecolor{ChongyangColor}{rgb}{0.0,0.6,0} 
\newcommand{\chongyang}[1]{{\color{ChongyangColor}[Chongyang: #1]}}
\definecolor{XingColor}{rgb}{0.0,0.8,0} 

\definecolor{FanColor}{rgb}{0.8,0,0.8}
\newcommand{\fan}[1]{{\color{FanColor}[Fan: #1]}}
\definecolor{OliverColor}{rgb}{0.54,0.18,0.88}
\newcommand{\oliver}[1]{{\color{OliverColor}[Oliver: #1]}}
\newcommand{\warning}[1]{{\it\color{red} #1}}
\newcommand{\toremove}[1]{{\it\color{red} (To remove) #1}}
\newcommand{\note}[1]{{\it\color{blue} #1}}
\newcommand{\nothing}[1]{}

\definecolor{AudioColor}{rgb}{0.56,0.34,0.62}

\definecolor{figred}{rgb}{1,0,0}
\definecolor{figgreen}{rgb}{0,0.6,0}
\definecolor{figblue}{rgb}{0,0,1}
\definecolor{figpink}{rgb}{1,0.63,0.63}

\ifthenelse{\equal{\final}{1}}
{
\renewcommand{\weiming}[1]{}
\renewcommand{\chongyang}[1]{}
\renewcommand{\yiping}[1]{}
\renewcommand{\huaiyu}[1]{}
\renewcommand{\fan}[1]{}
\renewcommand{\oliver}[1]{}
\renewcommand{\warning}[1]{}
\renewcommand{\toremove}[1]{}
\renewcommand{\note}[1]{}
}
{}

\renewcommand{\paragraph}[1]{\textbf{#1}}

\usepackage[accepted]{icml2019}

\icmltitlerunning{LGM-Net: Learning to Generate Matching Networks for Few-Shot Learning}

\begin{document}

\twocolumn[
\icmltitle{LGM-Net: Learning to Generate Matching Networks for Few-Shot Learning}



\icmlsetsymbol{equal}{*}

\begin{icmlauthorlist}
\icmlauthor{Huaiyu Li}{casia,ucas}
\icmlauthor{Weiming Dong}{casia}
\icmlauthor{Xing Mei}{am1}
\icmlauthor{Chongyang Ma}{am2}
\icmlauthor{Feiyue Huang}{te}
\icmlauthor{Bao-Gang Hu}{casia}
\end{icmlauthorlist}

\icmlaffiliation{casia}{National Laboratory of Pattern Recognition, Institute of Automation, Chinese Academy of Sciences, Beijing 100190, China}
\icmlaffiliation{ucas}{University of Chinese Academy of Sciences, Beijing 100049, China}
\icmlaffiliation{am1}{Snap Inc.}
\icmlaffiliation{am2}{Kwai Inc.}
\icmlaffiliation{te}{Youtu Lab, Tencent}

\icmlcorrespondingauthor{Weiming Dong}{weiming.dong@ia.ac.cn}

\icmlkeywords{Machine Learning, ICML}

\vskip 0.3in
]



\printAffiliationsAndNotice{}  

\begin{abstract}
In this work, we propose a novel meta-learning approach for few-shot classification, which learns transferable prior knowledge across tasks and directly produces network parameters for similar unseen tasks with training samples. 
Our approach, called LGM-Net, includes two key modules, namely, TargetNet and MetaNet. 
The TargetNet module is a neural network for solving a specific task and the MetaNet module aims at learning to generate functional weights for TargetNet by observing training samples. 
We also present an intertask normalization strategy for the training process to leverage common information shared across different tasks.
The experimental results on Omniglot and \textit{mini}ImageNet datasets demonstrate that LGM-Net can effectively adapt to similar unseen tasks and achieve competitive performance, and the results on synthetic datasets show that transferable prior knowledge is learned by the MetaNet module via mapping training data to functional weights. 
LGM-Net enables fast learning and adaptation since no further tuning steps are required compared to other meta-learning approaches.

\end{abstract}

\section{Introduction}
The ability to rapidly learn and generalize from a small number of examples is a critical characteristic of human intelligence, because humans can leverage the prior knowledge obtained from previous learning experience~\cite{2017building}.
Although current deep learning approaches have achieved significant success in many tasks~\cite{2012alexnet,2015fasterrcnn,2015fcn}, massive labeled data and excessive training time are still required, because each task is independently considered and the model parameters are learned from scratch without incorporating task-specific prior knowledge.  
How to extract prior knowledge and transfer them to unseen tasks with limited data has become active research areas in machine learning, such as transfer learning~\cite{2016tlsurvey}, metric learning~\cite{2015siamese}, and domain adaptation~\cite{2017fsada}.

In this work, we focus on few-shot learning, which aims at learning to recognize unseen categories from few labeled samples.
Recently, meta-learning~\cite{thrun2012learning} has emerged as a kind of promising approach to solve this problem.
A generic meta-learning framework usually contains a meta-level learner and a base-level learner~\cite{2016learning2learn, 2017metanets}. The base-level learner is designed for specific tasks, such as classification, regression, and neural network policy~\cite{2017maml}.
The meta-level learner aims to learn prior knowledge across different tasks.
The prior knowledge can be transferred to the base-level learner to help quickly adapt to similar unseen tasks.
For instance, \cite{2017meta-lstm} proposed to train the meta-level learner as the weight optimizer of a base-level learner.
The prior knowledge was embedded in how to update the base-level learner on a new task.
\cite{2017maml} proposed to learn a good initialization that can be quickly adapted to new tasks with a few updates.
The prior knowledge was learned and embedded in the good initialization.

In this work, we propose a novel meta-learning approach for few-shot learning. In contrast to previous methods ~\cite{2017metanets,2017meta-lstm,2016learning2learn}, our approach directly generates the functional weights of a network with limited training samples.
Our approach contains two key modules, namely, a TargetNet module (the base-level learner) and a MetaNet module \footnote{Our MetaNet module has the same abbreviation with, but is different from, Meta Networks~\cite{2017metanets} method.} (the meta-level learner).
The TargetNet module is designed for specific tasks. Traditionally, the parameters of this network are randomly initialized and adjusted by stochastic gradient descent (SGD) algorithms applied on training data. However, the parameters in TargetNet are generated by the MetaNet module conditioned on training samples.
The MetaNet contains two parts, namely, a \emph{task context encoder} designed to learn task context representation and a \emph{weight generator} that learns the conditional distribution of functional weights of TargetNet. 
We train the task context encoder on many tasks to encode training data of each task. The weight generator is trained to generate the weights of TargetNet for each task conditioned on encoded task representation.
The prior knowledge about how to generate functional weights using training data is learned by the MetaNet module, which is a new form of representing transferable prior knowledge.
We apply an intertask normalization(ITN) strategy which leverages information shared among different tasks to help training.

After training, MetaNet can generate the weights of TargetNet which can effectively generalize to similar unseen tasks. In the proposed approach, no further fine-tuning steps are necessary when meet unseen tasks because we directly produce functional weights.
Accordingly, rapid learning and adaptation on new tasks are achieved.
More specifically, we use matching networks~\cite{2016matchingnets} as the TargetNet structure and MetaNet learns to generate the weights for matching networks. Hence, we denote our approach as LGM-Net. 
Our contributions can be summarized as follows:
\begin{itemize}
    \item A novel meta-learning algorithm that trains MetaNet to generate the weights of TargetNet on the basis of training samples. 
    \item A simple and effective MetaNet module that encodes training samples and learns conditional distribution of TargetNet parameters.
\end{itemize}
We demonstrate the effectiveness of our approach on Omniglot and \textit{mini}ImageNet datasets. LGM-Net significantly outperforms many other meta-learning methods, especially for \textit{mini}ImageNet. We also conduct extensive experiments to show the mechanism of LGM-Net and analyze the generated weights on different tasks for validating our approach.

\section{Related Work}
Our approach aims at rapidly adapting to similar unseen tasks via generating weights using limited data. Many meta-learning methods are relevant to our work.

Matching networks~\cite{2016matchingnets} use an attention mechanism in the embedding space of training samples to predict classes for testing samples.
This model proposes an episodic training scheme where each episode is designed to mimic a few-shot task.
Several recent meta-learning approaches extend this episodic training idea.
\cite{2017meta-lstm} presented an LSTM-based meta-level learner to learn the exact learning rules which can be utilized to train a neural network classifier in a few-shot regime.
MAML\cite{2017maml} learns a good initialization of neural networks, which can be fine-tuned in a few gradient steps and effectively generalized on new unseen few-shot tasks.
Other approaches propose to learn a good initialization with an update function~\cite{2017meta-sgd} or without any update steps~\cite{2018reptile}. 
These meta learning methods represent the transferable prior knowledge in the good initialization or learned update functions.
In contrast to their approaches, our method represents prior knowledge in encoding the few-shot task samples and producing functional parameters of the TargetNet for each task without fine-tuning.

Another relevant direction involves the exploration of using one neural network to produce the parameters of another.
The conceptual framework of fast weights~\cite{1987fast,2016usingfast} is proposed to simulate synaptic dynamics at different time scales for storing temporary memories.
\cite{2017metanets} presented a meta- learning model that is equipped with external memory and utilizes meta-information to quickly parameterize both a meta-level learner and a base-level learner.
\cite{2018qiao} proposed to learn to predict the parameters of a top layer from the activations and adapt the pre-trained neural networks to novel categories.
\cite{2018dflwf} used an attention based weight generator to predict the weights for novel categories from pre-trained weights without forgetting previously learned categories.
\cite{2018leo} proposed to learn a data-dependent latent representation of model parameters and perform meta-learning in the latent space rather than in the parameter space as~\cite{2017maml}.
\cite{2018mela} proposed to learn a model code via a meta-recognition model and to construct parameters for a task-specific model by using a meta-generative model.

\section{Our Method}
\label{sec:method}
\begin{figure*}
	\centering
	\includegraphics[width=0.9\textwidth]{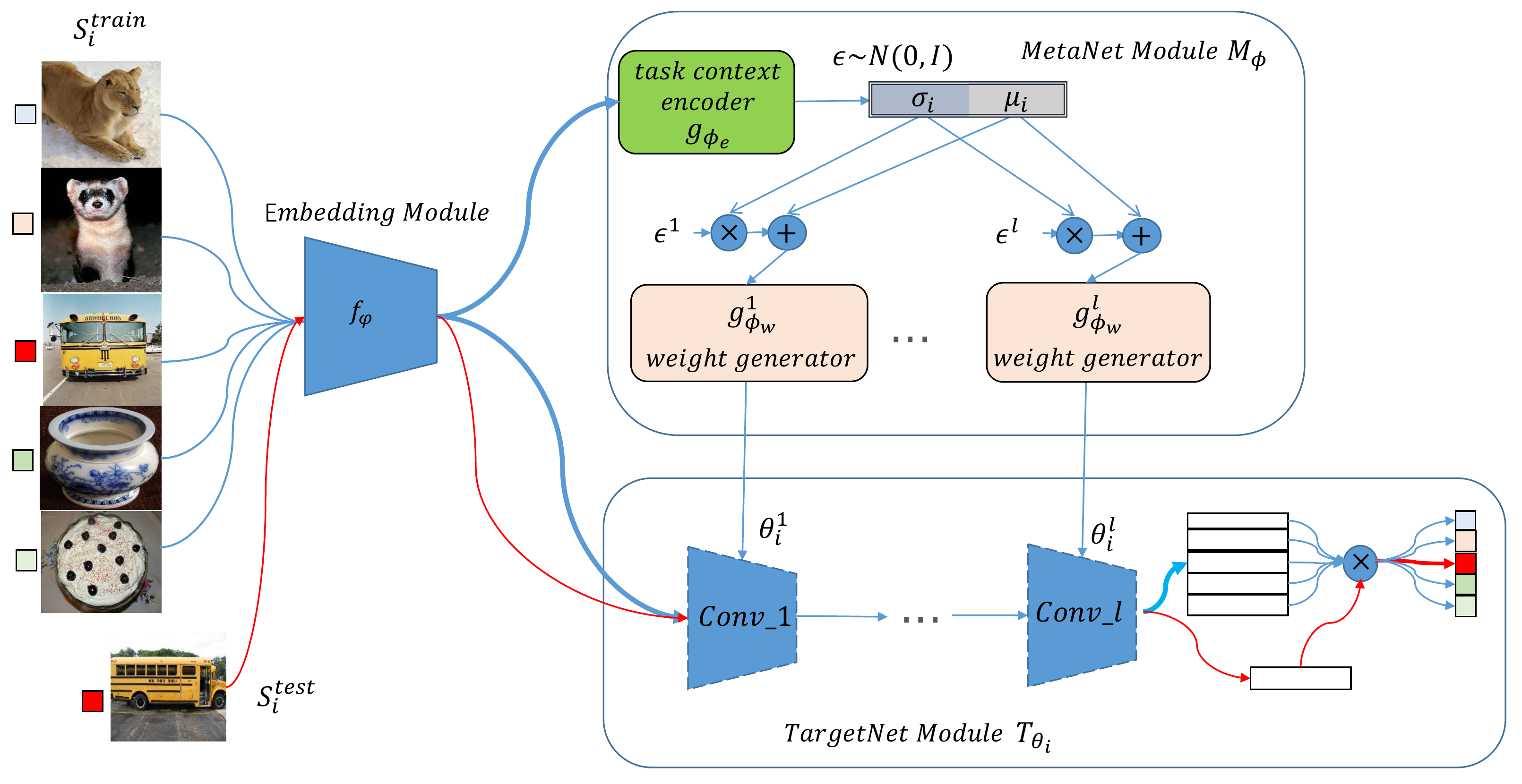}
	\vspace{-3pt}
	\caption{The architecture of our LGM-Net for few-shot learning on 5-way 1-shot classification problems.
}
	\label{fig:arch}
\end{figure*}

\subsection{Problem Formulation}
We define an $N$-way $K$-shot problem using the episodic formulation from \cite{2016matchingnets}. 
Formally, we have three datasets, i.e., a meta training dataset $D^{meta\text{-}train}$, a meta validation dataset $D^{meta\text{-}val}$, and a meta test dataset $D^{meta\text{-}test}$.
Each dataset contains a disjoint set of target classes. 
For each dataset, we can construct a task distribution $p(\mathcal{T})$ of $N$-way $K$-shot tasks.
Each task instance $\mathcal{T}_i \sim p(\mathcal{T})$ consists of a training set $S_i^{train}$ and a test set $S_i^{test}$.
The training set $S_i^{train}$ contains $N$ classes randomly selected from a meta dataset and $K$ samples for each class.
The test set $S_i^{test}$ contains unseen samples for classes in $S_i^{train}$ and provides an estimation of generalization performance on the $N$ classes for task $\mathcal{T}_i$.
We use the meta training dataset to train our model and the meta validation dataset for model selection.
The meta test dataset is only used to evaluate the model generalization on unseen tasks.

\subsection{Preliminary}
The functionality of a deep neural network (DNN) depends on its weights and architecture. When the architecture of a neural network is fixed, the weights determine its functionality.
For instance, the AlexNet architecture~\cite{2012alexnet} can be trained to classify $100$ types of animals or 100 categories of plants. However, the two AlexNets have different weights\nothing{ with different functionality}.
A network has no functionality when the weights are randomly initialized.
In this work, we refer to the weights that embody networks abilities as \emph{functional weights}.

When we train a DNN for a specific task, we usually use gradient descent algorithms~\cite{2016sgdoverview} to optimize randomly initialized weights on the training data for a finite number of iterations.
Since the loss landscape~\cite{2017visualizing} of neural networks is extremely complicated, the optimization process often converges to different local optima or saddle points.
For test samples, these points in the weight space usually present similar generalization ability~\cite{2017porcupine}.
Therefore, we can consider these functional weights as a distribution over training data.
We design the meta-learning approach that can learn the conditional functional weight distribution over training data and directly produce functional weights for DNNs with limited training samples.
\begin{algorithm}[t]
	\caption{The training algorithm of LGM-Net for $N$-way $K$-shot problems}
	\label{alg:metanet_training}
	\begin{algorithmic}
		\STATE {\bfseries Required:} Meta training dataset $D^{meta\text{-}train}$
		\STATE {\bfseries Required:} MetaNet $\mathit{M}$ with parameters $\phi$, TargetNet computational structure $\mathit{T}$ with parameter placeholder $\theta$.
		\STATE{Randomly initialize $\phi$}
		\WHILE{not converged}
		\STATE{Sample a $N$-way $K$-shot task batch $\mathcal{T}^{batch}$ from $D^{meta\text{-}train}$}
		\FOR{all the task instances in a batch}
		\STATE{Divide a task instance as $(S_i^{train}, S_i^{test})=\mathcal{T}_i$}
		\STATE{Sample a functional weights point $\hat{\theta}$ for TargetNet from $\mathit{M}(S_i^{train})$}
		\STATE{Assign generated weights $\hat{\theta}$ to TargetNet placeholder weights $\theta$}
		\STATE{Compute TargetNet test loss for this task on $S_i^{test}$ as $\mathcal{L}_{\mathcal{T}_i}$}
		\ENDFOR
		\STATE{Compute batch loss $\mathcal{L}_{\mathcal{T}^{batch}} = \sum_{\mathcal{T}_i} \mathcal{L}_{\mathcal{T}_i}$}
		\STATE{Update $\phi$ using $\nabla_{\phi} \mathcal{L}_{\mathcal{T}^{batch}}$}
		\ENDWHILE
	\end{algorithmic}
\end{algorithm}

\subsection{Methodology}

As illustrated in Figure~\ref{fig:arch}, our LGM-Net architecture consists of two key modules, namely, a TargetNet module and a MetaNet module.
The training procedure is summarized in Algorithm~\ref{alg:metanet_training}. 
We first get a batch of task $\mathcal{T}^{batch}$ given the meta training dataset.
For each task instance $\mathcal{T}_i$, the MetaNet module generate a functional weight point $\hat{\theta} = \mathit{M}(S_i^{train})$ for the TargetNet conditioned on training set. Then, the TargetNet assigned with generated weights can infer the matching probability scores for test samples. The classification loss is simultaneously computed. Finally, for each task in a batch, we accumulate the losses and compute the gradient updates for the parameters in MetaNet module.
For high dimensional input data, a learnable embedding module $f_{\varphi}$ is used to extract low dimensional features as inputs for the two modules. In this way, the amount of parameters of the entire model can be reduced.

\subsection{MetaNet Module}
Our MetaNet module consists of a task context encoder (Section~\ref{sec:task_context_encoder}) and a conditional weight generator (Section~\ref{sec:conditional_weight_generator}).

\subsubsection{Task Context Encoder}
\label{sec:task_context_encoder}
The task context encoder aims to encode all the training samples of a task and generate a feature representation of the task with fixed size. The task context features should satisfy the following properties to appropriately represent a task: enough distinctions between different tasks, sufficient similarities between similar tasks, and insensitivity to the number and order of samples in a certain task.
More importantly, the encoding operation should be differentiable.
In Matching networks~\cite{2016matchingnets}, BiLSTMs~\cite{1997lstm} are used to extract fully contextual embeddings.
However, using BiLSTMs to encode task contexts in our case makes training and converging difficult. 
We compute the summary statistics of training set of each task without any supervision for simplicity and effectiveness in accordance with the Neural Statistician~\cite{2017neuralstatistician}.
The task context features are reparameterized as a conditional multivariate Gaussian distribution with a diagonal covariance.
Specifically, given an $N$-way $K$-shot task $\mathcal{T}_i$ with a training set $S_i^{train} = \big\{ (x_i^{n,k}, y_i^{n,k}) | k=1...K, n=1...N \big\} $, we formulate the sampling of the task context as follows:
\begin{equation}
\mathbf{\mu}_i, \mathbf{\sigma}_i = \frac{1}{NK}\sum_{n=1}^{N} \sum_{k=1}^{K} g_{\phi_{e}}(x_i^{n,k}),
\end{equation}
\begin{equation}
\mathbf{c}_i \sim q(\mathbf{c}_i | S_i^{train}) = \mathcal{N}\big(\mathbf{\mu}_i, \mathrm{diag}({\mathbf{\sigma}_i}^2)\big),
\end{equation}
where $g_{\phi_{e}}$ is the task context encoder, $q$ denotes a conditional probability distribution of task context features and $\mathbf{c}_i$ denotes a sample of task context features for $\mathcal{T}_i$.

\subsubsection{Conditional Weight Generator}
\label{sec:conditional_weight_generator}

The weight generator is trained to generate the functional weights for TargetNet conditioned on task context features.
For each layer of the TargetNet, we construct a conditional single layer perceptron as the generator to produce weights as follows:
\begin{equation}
	\theta_i^l = g_{\phi_{w}}^l(\mathbf{c}_i),
\end{equation}
where $g_{\phi_{w}}^l$ is the weight generator for $l$-th layer. 

We apply weight normalization (WN) to constrain the weight scale for facilitating the training process, but remove learnable parameters compared with the original WN method~\cite{2016weightnorm}.
For the generated weights of a convolution layer, the L2 normalization is applied to each kernel rather than the entire convolution weights.
For the generated weights of a fully connected layer, the L2 normalization is applied to each hyperplane weights, which can be formulated as:
\begin{equation}
\theta_{i,j}^l = \frac{\theta_{i,j}^l}{||\theta_{i,j}^l||_2},
\end{equation}
where $\theta_{i,j}^l$ is the $j$-th kernel or hyperplane weight of generated weights of $l$-th layer on $i$-th task.
In practice, the WN helps prevent generating large scale features and stabilize the training process.

\subsection{TargetNet Module}
We use matching networks as the architecture of TargetNet for few-shot classifcation problems.
The functional weights of TargetNet are generated by MetaNet based on training samples.
There are many newly designed parametric layers in neural networks, such as parametric ReLU~\cite{2015prelu} and batch normalization~\cite{2015bn}.
These layers contain learnable parameters and aim to stabilize the training of DNNs.
Therefore, we only consider generating convolutional kernels, bias, and fully connected weights. 
We denote the TargetNet for an $N$-way $K$-shot task $\mathcal{T}_i$ as $T_{\theta_i}$ and the corresponding test set of $\mathcal{T}_i$ as $S_i^{test} = \left\lbrace(\hat{x}_i, \hat{y}_i) \right\rbrace$ in accordance with the previous description.
The final embeddings of the training samples $T_{\theta_i}(x_i^{n,k})$ and test samples $T_{\theta_i}(\hat{x}_i)$ are obtained through TargetNet.
The probability attention kernel\cite{2016matchingnets} is computed as follows:
\begin{equation}
a(\hat{x}, x_i) = \frac{e^{d(T_{\theta_i}(\hat{x}_i), T_{\theta_i}(x_i^{n,k})))}}{\sum_{n=1}^{N} \sum_{k=1}^{K}e^{d(T_{\theta_i}(\hat{x}_i), T_{\theta_i}(x_i^{n, k})))}},
\end{equation}
where $d(T_{\theta_i}(\hat{x}_i), T_{\theta_i}(x_i^{n,k}))$ is the cosine distance between test and training sample embeddings.
The probability attention is then used to obtain the final probability score of the test sample, which is formulated as:
\begin{equation}
\mathbf{\hat{p}}_i = \sum_{n=1}^{N} \sum_{k=1}^{K}a(\hat{x}_i, x_i^{n,k})\mathbf{y}_i^{n,k}.
\end{equation}
We adopt cross-entropy loss to construct the final objective function between the predicted probability and the groundtruth:
\begin{equation}
\mathcal{L}_{\mathcal{T}_i} = H(\mathbf{\hat{y}}_i, \mathbf{\hat{p}}_i).
\end{equation}

\subsection{Intertask Normalization}
Previous meta-learning methods usually consider each task independently.
However, similar tasks should share some useful information with each other, which can help meta-level learner to learn additional common prior knowledge. 
We propose an intertask normalization (ITN) strategy to make the tasks interact with each other in a batch of tasks.
In practice, we directly apply batch normalization~\cite{2015bn} on the embedding module and task context encoder.
The normalization is applied to all training samples of a task batch, rather than just to samples of each individual task.
The accumulated mean, variance and learned scale and shift parameters in BN incorporate the statistical information shared among tasks.
During a testing phase, we independently apply the trained model on each individual unseen task.

\section{Experiments}
In this section, we conduct several experiments to evaluate and verify our proposed method.
In Section~\ref{sec:toy_example}, we apply LGM-Net on four synthetic datasets and visualize the decision boundary of generated TargetNet on unseen tasks\nothing{, to show the intuition of our approach}.
In Section~\ref{sec:few_shot}, we perform few-shot image classification and compare with state-of-the-art methods on Omniglot and \emph{mini}ImageNet datasets.
In Section~\ref{sec:ablation}, we conduct ablation study to evaluate each component of our approach.
In Section~\ref{sec:exploring}, we visualize the distribution of generated weights on different unseen tasks.
We implement our algorithm and conduct the experiments using TensorFlow~\cite{2016tensorflow}.
Our source code is available online\footnote{https://github.com/likesiwell/LGM-Net/}.

\begin{figure}
	\centering
	\begin{subfigure}[b]{0.23\textwidth}
		\includegraphics[width=\textwidth, trim=60px 45px 60px 45px, clip]{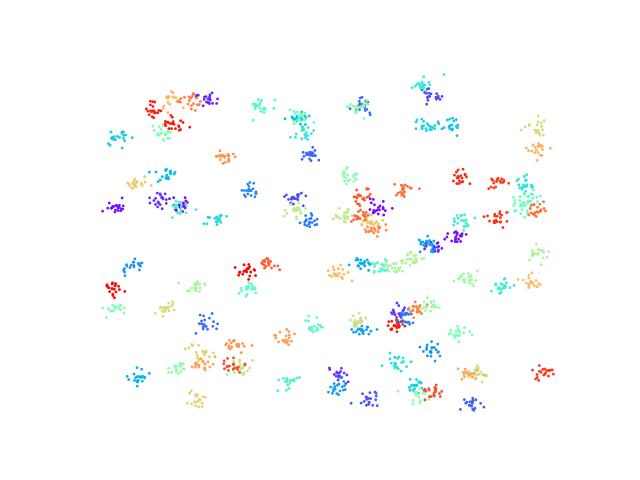}
		\caption{Blobs}
	\end{subfigure}
	\begin{subfigure}[b]{0.23\textwidth}
		\includegraphics[width=\textwidth, trim=60px 45px 60px 45px, clip]{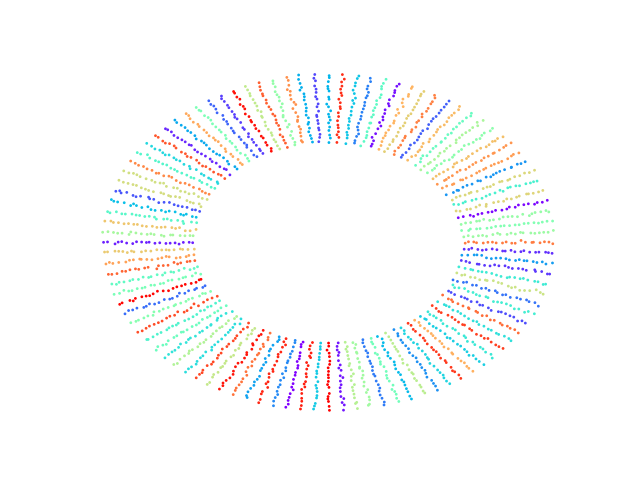}
		\caption{Lines}
	\end{subfigure}
	\begin{subfigure}[b]{0.23\textwidth}
		\includegraphics[width=\textwidth, trim=60px 45px 60px 45px, clip]{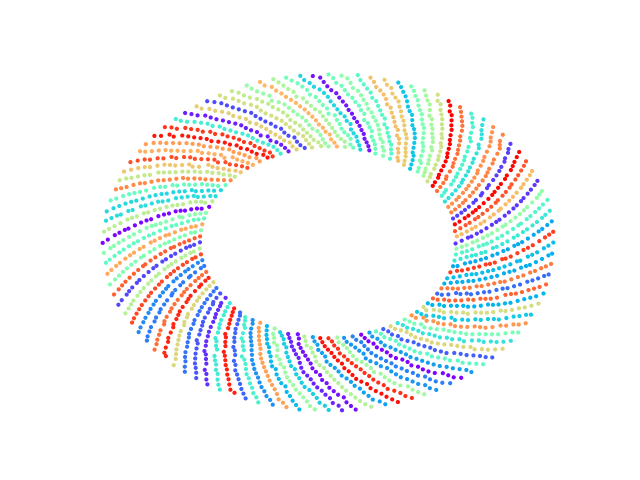}
		\caption{Spirals}
	\end{subfigure}
	\begin{subfigure}[b]{0.23\textwidth}
		\includegraphics[width=\textwidth, trim=60px 45px 60px 45px, clip]{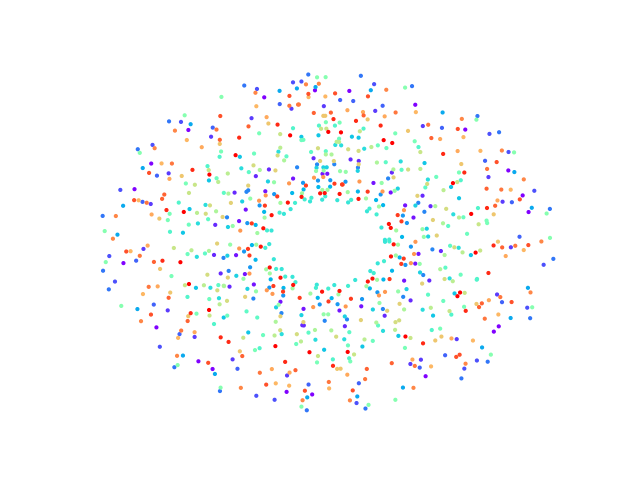}
		\caption{Circles}
	\end{subfigure}
	\caption{Visualization of four synthetic datasets. The blobs, lines, spirals, and circles in the same color belong to the same category. Each dataset contains 100 categories with 20 samples per category. This figure is best viewed in electronic version.}
	\label{fig:toy}
\end{figure}

\subsection{Results on Synthetic Datasets}
\label{sec:toy_example}

\begin{figure}[t]
\centering
\includegraphics[width=0.32\linewidth]{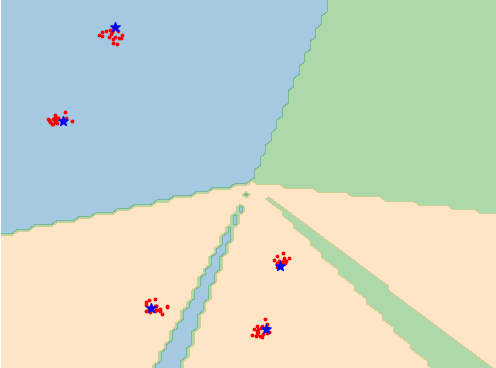}
\includegraphics[width=0.32\linewidth]{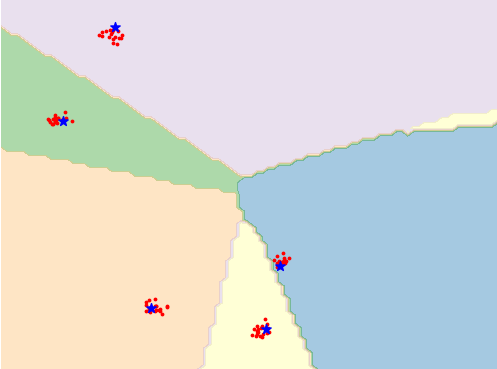}
\includegraphics[width=0.32\linewidth]{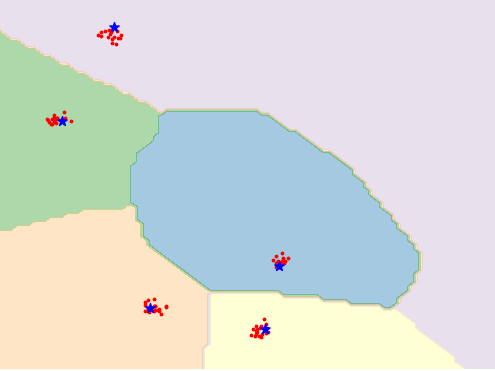}

\includegraphics[width=0.32\linewidth]{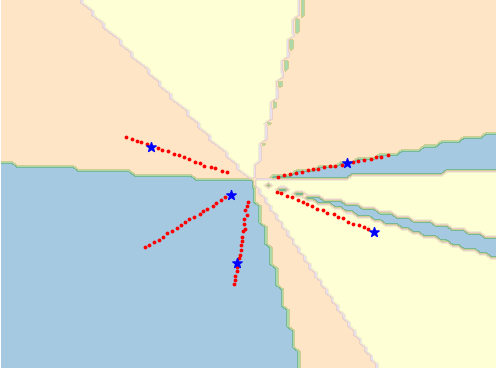}
\includegraphics[width=0.32\linewidth]{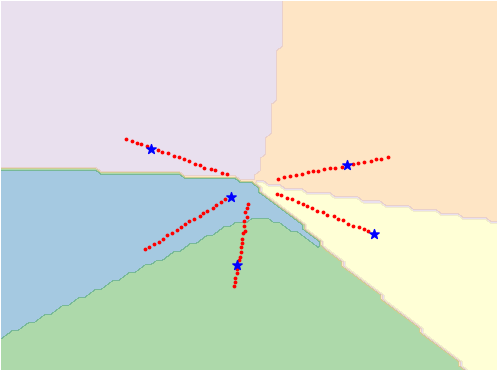}
\includegraphics[width=0.32\linewidth]{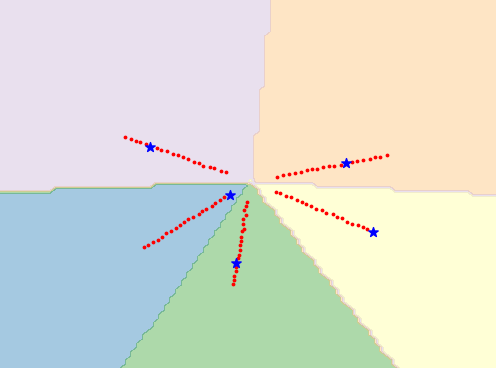}

\includegraphics[width=0.32\linewidth]{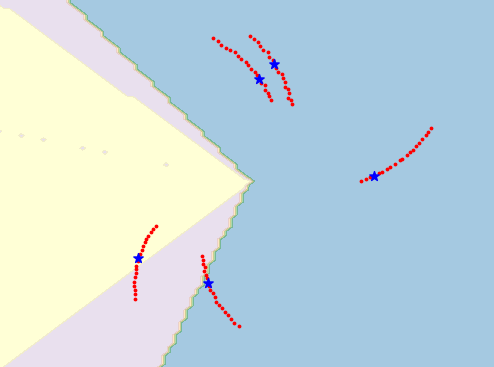}
\includegraphics[width=0.32\linewidth]{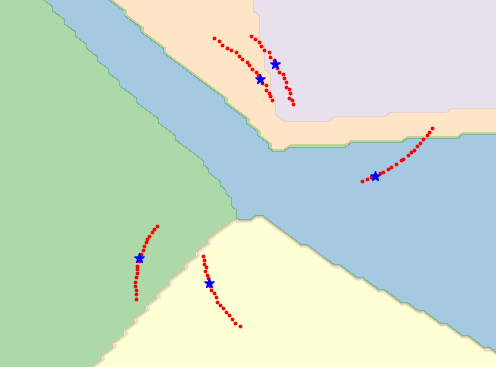}
\includegraphics[width=0.32\linewidth]{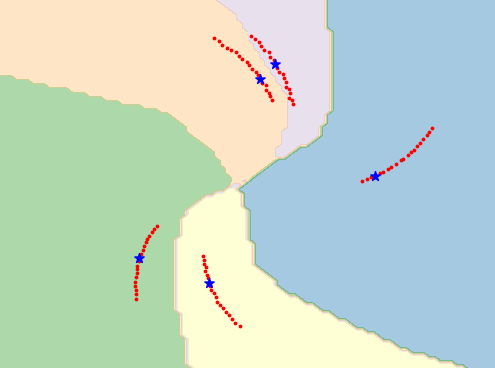}

\includegraphics[width=0.32\linewidth]{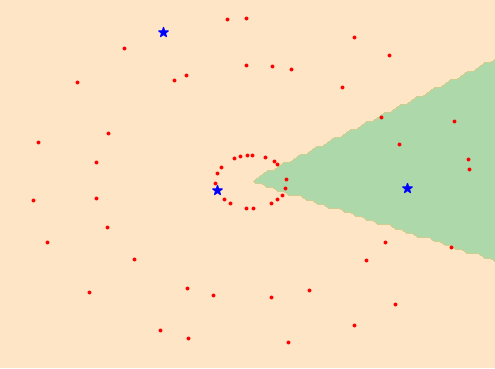}
\includegraphics[width=0.32\linewidth]{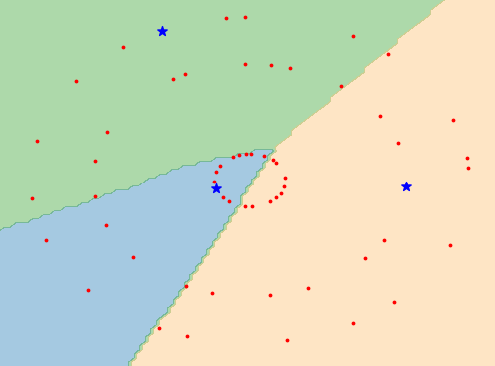}
\includegraphics[width=0.32\linewidth]{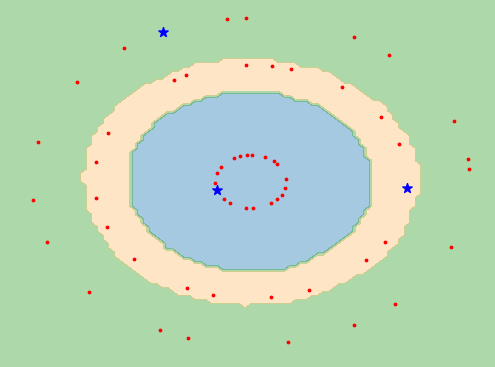}
\caption{The decision boundaries of TargetNet on four synthetic datasets (from top to bottom: \emph{Blobs}, \emph{Lines}, \emph{Spirals}, and \emph{Circles}).
The blue star points denote the training set of a selected task $\mathcal{T}_i$ from the meta testing dataset.
The red points are the test samples of $\mathcal{T}_i$.
In each row, from left to right, we show the decision boundaries of TargetNet from randomly initialized weights, directly trained weights using Adam optimization, as well as the weights generated by our MetaNet module.
}
\label{fig:boundary}
\end{figure}

To begin with, we consider an experiment on synthetic datasets to show the intuition and testify the effectiveness of our method. 
As shown in Figure \ref{fig:toy}, we have designed four 2D synthetic datasets, i.e., \emph{Blobs}, \emph{Lines}, \emph{Spirals}, and \emph{Circles}.
To formalize these datasets in the context of few-shot learning, we consider blobs in different positions, lines along different angles, different spirals arms, as well as concentric circles with different radius as different categories in each dataset.
The data points of the same color belong to the same category.
Each dataset contains $100$ categories with $20$ samples per category.
We randomly select $80$ categories as meta training dataset and the remaining $20$ categories as meta test dataset.

We can directly feed the training set of a task into MetaNet and generate the weights of TargetNet because the synthetic datasets are 2D.
In our experiments, the task context encoder consists of a multilayer perceptron (MLP) with two hidden layers of $8$ units and a ReLU activation function.
The computational structure of TargetNet includes an MLP with three hidden layers of $16$, $12$, and $8$ units and ReLU activation layers.
For better visualization, we train the \emph{Blobs}, \emph{Lines}, and \emph{Spirals} datasets on 5-way 1-shot learning tasks but train the \emph{Circles} dataset on 3-way 1-shot learning tasks. 

Figure~\ref{fig:boundary} shows the decision boundaries of TargetNet on the aforementioned four synthetic examples using three different approaches.
The learning difficulties gradually increase from top to bottom in the figure\nothing{ \emph{Blobs}, \emph{Lines} to \emph{Spirals} and \emph{Circles}}.
From left to right, in the first column, we plot the decision boundaries of TargetNet with random initialized weights for an unseen task $\mathcal{T}_i$ of each synthetic data. The decision boundaries are in disorder.
In the second column, we directly train TargetNet on the training set of task $\mathcal{T}_1$ using Adam optimization~\cite{2014adam}.
The results show that, the directly trained TargetNet cannot effectively generalize to the test samples well even though it can correctly classify the training samples.
Most notably, the decision boundaries for the task from \emph{Circles} show that the directly trained TargetNet misclassifies most of the training samples.
This fact is due to the traditional gradient descent algorithms that only consider training samples and contain no any prior knowledge about the specific task.
These algorithms usually overfit the model with limited training samples.
In the third column, we show the decision boundaries of TargetNet with generated weights from the learned MetaNet module based on $\mathcal{T}_i$.
The TargetNet using generated weights can effectively generalize on test samples on the selected task.
Our approach for all the four datasets can achieve over $99\%$ accuracy on unseen tasks. 

The experiments on synthetic examples demonstrate the learning mechanism of LGM-Net.
The decision boundaries on different dataset scenarios show that the MetaNet module learns to understand different task scenarios.
The MetaNet can directly generate functional weights for the TargetNet for a similar unseen task.
However, the TargetNet with directly trained weights fails to generalize well, because the SGD algorithms do not contain any prior knowledge for learning similar scenarios.
The comparison results imply that the MetaNet learns the transferable prior knowledge about how to solve tasks for a certain scenario.
Even when the unseen task contains few examples, the MetaNet can still generate effective functional weights for TargetNet. 

\subsection{Few-shot Classification Results}
\label{sec:few_shot}
We conduct few-shot classification experiments on two commonly used benchmarks, namely, Omniglot~\cite{2011omniglot} and \textit{mini}ImageNet~\cite{2016matchingnets}. 
Adam optimization~\cite{2014adam} is applied for training, with an initial learning rate of $10^{-3}$ which is reduced by $10\%$ every $1500$ batches.
The models are trained end-to-end from scratch without any additional dataset.

\begin{table*}[t]
	\caption{Mean accuracy of our LGM-Net and state-of-the-art methods on Omniglot dataset.
	}
	\label{tab:omni}
	\small
	\centering
	\begin{tabular}{lcccc}
		\toprule
		\bf Model & \bf 5-way 1-shot & \bf 5-way 5-shot & \bf 20-way 1-shot & \bf 20-way 5-shot \\
		\midrule
		Siamese Net \cite{2015siamese} & 97.3\% & 98.4\% & 88.1\% & 97.0\% \\
		Neural Statistician \cite{2017neuralstatistician} & 98.1\% & 99.5\% & 93.2\% & 98.1\% \\
		Meta Nets \cite{2017metanets} &  99.0\% & - & 97.0\% & - \\
		Prototypical Nets \cite{2017prototypicalnets} & 98.8\%  & 99.7\% & 96.0\% & 98.9\% \\
		MAML \cite{2017maml} & 98.7\% & \bf 99.9\% & 95.8\% & 98.9\% \\
		Meta-SGD \cite{2017meta-sgd} & 99.5\% & \bf 99.9\% & 95.9\% & 99.0\% \\
		Relation Net \cite{2017learn2compare} & \bf 99.6\% & 99.8\% & \bf 97.6\% & \bf 99.1\% \\
		Matching networks \cite{2016matchingnets} & 98.1\% & 98.9\% & 93.8\% & 98.5\% \\
		\midrule
		LGM-Net (Ours) & 99.0\% & 99.4\% & 96.5\% & 98.5\% \\	
		\bottomrule
	\end{tabular}
\end{table*}

\subsubsection{Results on Omniglot Dataset}
The Omniglot dataset consists of $1623$ characters (classes) with $20$ samples for each class from 50 different alphabets.
Following \cite{2016matchingnets,2017prototypicalnets}, we randomly select $1200$ classes as meta training dataset and use the remaining $423$ classes as meta test dataset.
The input images are resized to a resolution of $28\times28$ and augmented by random rotations of $90$, $180$, or $270$ degrees.

In this experiment, we follow the architecture used by \cite{2016matchingnets} to construct our modules.
The embedding module has three $3\times3$ convolutional layers with $64$ filters, followed by batch normalization, a ReLU activation, and $2\times2$ max-pooling. The TargetNet consists of two $3\times3$ convolutional layers with $64$ filters which are generated by MetaNet. The task context encoder contains two $3\times3$ convolutional layers with $64$ filters. 
During the test phase, we randomly select $8000$ N-way K-shot tasks from meta test dataset for evaluation purpose.
As shown in Table~\ref{tab:omni}, we achieve comparable performance against state-of-the-art few-shot learning methods under different experiment settings.
Since the performance scores are almost saturated on this dataset, the experimental results can still demonstrate the validity of our approach.
 
\begin{table*}[t]
	\caption{Mean accuracy $\pm$ 95\% confidence intervals of our LGM-Net and state-of-the-art methods on \textit{mini}ImageNet dataset.
	}
	\label{tab:mini}
	\small
	\centering
	\begin{tabular}{lcccc}
		\toprule
		\bf Model & \bf 5-way 1-shot & \bf 5-way 5-shot & \bf 20-way 1-shot  \\
		\midrule
		Matching networks \cite{2016matchingnets} & 43.56$\pm$0.84\% & 55.31$\pm$0.73\% & 17.31$\pm$0.22\%  \\
		Meta-LSTM \cite{2017meta-lstm} & 43.44$\pm$0.77\% & 60.60$\pm$0.71\% & 16.70$\pm$0.23\% \\
		MetaNet \cite{2017metanets} &  49.21$\pm$0.96\% & - & - \\
		Prototypical Nets \cite{2017prototypicalnets} & 49.42$\pm$0.78\% & 68.20$\pm$0.66\% &  \\
		MAML \cite{2017maml} & 48.70$\pm$1.84\% & 63.11$\pm$0.92\% & 16.49$\pm$0.58\% \\
		Meta-SGD \cite{2017meta-sgd} & 50.47$\pm$1.87\% & 64.03$\pm$0.94\% & 17.56$\pm$0.64\% \\
		Relation Net \cite{2017learn2compare} & 51.38$\pm$0.82\% & 67.07$\pm$0.69\% & - \\
		REPTILE \cite{2018reptile} & 49.97$\pm$0.32\% & 65.99 $\pm$ 0.58\% & - \\
		SNAIL \cite{2018snail} & 55.71$\pm$0.99\% & 65.99 $\pm$ 0.58\% & - \\
		\cite{2018dflwf} & 56.20$\pm$0.86\% &  73.00 $\pm$ 0.64\% & - \\
		LEO\cite{2018leo} & 61.76$\pm$0.08\% & \bf 77.59$\pm$ 0.12\% & - \\
		\midrule
		LGM-Net (Ours) & \bf 69.13$\pm$0.35\% & \bf 71.18$\pm$0.68\% & \bf 26.14$\pm$0.34\% \\		
		\bottomrule
	\end{tabular}
\end{table*}

\subsubsection{Results on \textit{mini}ImageNet Dataset}

The \textit{mini}ImageNet dataset, originally proposed by~\cite{2016matchingnets}, consists of 60,000 images from 100 selected ImageNet classes, each having 600 examples. 
We follow the split introduced by~\cite{2017meta-lstm}, with $64$, $16$, and $20$ classes for training, validation, and test, respectively.
All the images are resized to a resolution of $84\times84$ and are augmented by random rotation of $-45$, $-22.5$, $22.5$, or $45$ degrees.

On the \textit{mini}ImageNet benchmark, the results reported by different methods are obtained in different network configurations.
Furthermore, different methods may require network configurations.
Networks with deeper layers or more learnable parameters usually obtain better performance. However, for some methods such as~\cite{2017maml,2017meta-lstm}, they restrict the number of filters to alleviate overfitting.
We construct the network architecture with six convolutional layers from image inputs to outputs, to make a relatively fair comparison with these methods.
Hence, the embedding module has four $3\times3$ convolutional layers with $64$ filters, followed by batch normalization, a ReLU nonlinearity, and $2\times2$ max-pooling.
The TargetNet consists of two $3\times3$ convolutional layers with $64$ filters without BN and a global average pooling is appended at last.
We compare with other related methods in similar network settings.
The results are shown in Table~\ref{tab:mini}.
Our LGM-Net achieves state-of-the-art performance and significantly improves by up to $8\%$ on 5-way 1-shot learning problem. The success of our approach lies in the transferrable prior experience learned by the MetaNet.
In contrast to the baseline matching networks in which the weights are fixed for unseen tasks, the weights in our TargetNet are dynamically generated and the transferable prior knowledge is quickly adapted to new tasks.
Furthermore, compared to alternative methods such as MAML~\cite{2017maml}, Meta-LSTM~\cite{2017meta-lstm}, and Meta-SGD~\cite{2017meta-sgd}, our representation of transferable prior knowledge in the way of generating functional weights is more effective than learning a good initialization or a parameter optimizer.

\subsection{Ablation Study}
\label{sec:ablation}
We perform an ablation study with detailed results in Table~\ref{tab:ablation} to evaluate the effects of each component in our algorithm. 
To ensure a fair comparison, we reimplement a matching networks with six $3\times3$ convolution blocks with $64$ filters as the baseline.
Additionally, our LGM-Net has the same computational structure, but the weights of the last two layers are generated by MetaNet.

We compare matching networks and LGM-Net both with and without using ITN.
As shown in Table~\ref{tab:ablation}, using ITN significantly improves the performance.
If we train LGM-Net without task context encoder (TCE), then it is equivalent to the matching Network whose last two layers are based on a weight generator of a random prior.
No evident difference is found in the test performance.
However, the LGM-Nets with TCE show a noticeable improvement.
This fact indicates that TCE helps generate better weights for TargetNet, rather than just increasing the learning parameters in the entire model. 
If we train LGM-Net without WN, then it slightly decreases the performance.
We have found that without using WN, the features in TargetNet are usually of large scale which will have negative impact on the training process.
If we train LGM-Net directly using task context feature to generate the weights rather than formulating the weight generator as sampling from a reparameterized multivariate Gaussian distribution, then the performance will be worse, which shows that randomness helps improve the performance. 
Moreover, we can use the weight generator to directly generate several weight points for a single task. These generated weights can directly form an ensemble model with a max voting algorithm. However, this approach leads to no evident improvement and only reduces variance. 

\begin{table}[t]
	\caption{Abalation Study on the \textit{mini}ImageNet dataset.
	}
	\label{tab:ablation}
	\small
	\centering
	\begin{tabular}{lcccc}
		\toprule
		\bf Model & \bf 5-way 1-shot & \bf 5-way 5-shot   \\
		\midrule
		matching networks(w/o ITN) & 44.98$\pm$0.88\% & 55.74$\pm$0.75\% \\ 
		matching networks(w ITN) & 57.42$\pm$0.68\% & 59.65$\pm$0.83\%\\  
		LGM-Net (w/o TCE) & 57.91$\pm$0.41\% & 59.89$\pm$0.48\% \\  
		LGM-Net (w/o ITN) & 65.42$\pm$0.65\% & 67.93$\pm$0.56\% \\ 
		LGM-Net (w/o randomness) & 67.25$\pm$0.42\% & 69.68$\pm$0.55\%\\
		LGM-Net (w/o WN) & 68.85$\pm$0.41\% & 69.97$\pm$0.45\% \\
		\midrule
		LGM-Net (Plain) & 69.13$\pm$ 0.35\% & 71.18$\pm$0.68\% \\		
		LGM-Net (Ensemble) & 69.15$\pm$ 0.33\% & 71.18$\pm$0.64\%  \\
		\bottomrule
	\end{tabular}
\end{table}

\subsection{Exploring Generated Weight Distribution}
\label{sec:exploring}
\begin{figure}
	\centering
	\includegraphics[width=0.8\linewidth, trim=70px 40px 70px 50px, clip]{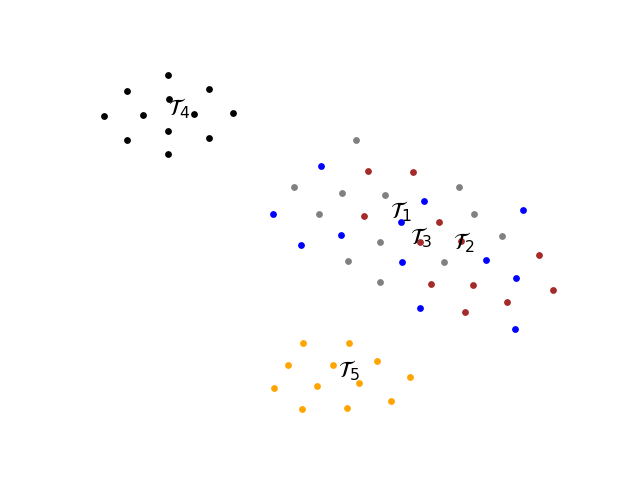}
	\vspace{-10pt}
	\caption{Distributions of generated weights for different tasks.}
	\label{fig:vis}
\end{figure}

In this experiment, the property of the MetaNet module is explored.
We select five 5-way 1-shot learning tasks from the meta test dataset.
For these tasks, $\mathcal{T}_1$, $\mathcal{T}_2$, and $\mathcal{T}_3$ contain the same training samples but in different orders, whereas $\mathcal{T}_4$ and $\mathcal{T}_5$ are another two different tasks.
Figure~\ref{fig:vis} shows the t-SNE~\cite{2008tsne} visualization of the generated functional weight points from the MetaNet conditioned on the selected tasks. For each task, we generate $12$ functional weight points from the MetaNet, which are marked in the same color.
As shown in Figure~\ref{fig:vis}, the functional weight points from the same task are clustered together.
Similar generalization ability is exhibited on the test set for the corresponding task. 
The weight point clusters of $\mathcal{T}_1$, $\mathcal{T}_2$, and $\mathcal{T}_3$ are overlapped together but far away from the clusters of $\mathcal{T}_4$ and $\mathcal{T}_5$.
This result indicates that the generated functional weights from MetaNet are not influenced by the order of training data, which is a desired property of our task context encoder.
Furthermore, different tasks have distinct distributions of functional weights.

\section{Conclusions}
Inspired by the ability of human beings to quickly learn from one task and adapt to a similar unseen task, we propose a novel meta learning approach, namely LGM-Net, for few-shot learning.
Our key idea is to learn a weight generation network across a large amount of tasks to produce functional weights for TargetNet based on the training data.
We choose Matching Nets as the computational structure of TargetNet and design a task context encoder and a weight generator as MetaNet. Compared with recent meta learning algorithms, LGM-Net is simpler and more efficient since it neither contains complicated structures nor needs further fine-tuning. It can learn transferable prior knowledge and enables fast adaptation to new tasks with limited data.

If we leave out the MetaNet module in LGM-Net, the computation pipeline is similar to convolutional siamese nets~\cite{2015siamese} and matching networks~\cite{2016matchingnets}.
The main difference among the three models lies in the learned parameters of the non-linear mapping function from the input to the embeddings. The siamese nets are trained on paired inputs with contrastive loss.
The matching networks are trained on few-shot tasks with attentional metric loss. The learned parameters in their nonlinear mapping functions remain unchanged even for different unseen tasks and therefore are difficult to adapt.
However, the parameters in our LGM-Net are adaptive to different task samples.
Hence, we can get better mapping function for new tasks.

The proposed approach has achieved significant improvement on 5-way 1-shot learning tasks on \textit{mini}ImageNet dataset.
However, on 5-way 5-shot learning problems, the performance improvement over 5-way 1-shot is not as obvious as other methods.
This outcome is due to the straightforward design of task context encoder which computes the statistical mean of training sample features as the task representation.
Although the task context encoder is simple and effective on 1-shot learning, it may not provide sufficient information with few-shot samples, thereby leading to limited improvement on few-shot learning tasks. 
Hence, designing an effective task context encoder will be one of future work.

Finally, LGM-Net, like other meta learning approaches, needs to be advanced towards explainable AI (XAI)~\cite{tickle1998the,doshi-velez2017towards,murdoch2019interpretable} or transparent AI~\cite{hu2007add}.
These meta learning methods extract prior knowledge, embed it into meta-level learner, and help base-level learner for solving novel tasks.
However, one critical question may arise, i.e., how can we interpret the prior knowledge embedded in meta-level learners and can we believe that it will guarantee better performance than conventional methods?
Therefore, it is necessary to design more explainable meta-level learners which can represent prior knowledge in more transparent forms and help us understand how to leverage the embedded prior knowledge to solve new tasks.
The related challenges can be another interesting direction for future work.

\section*{Acknowledgements}
\label{sec:ack}

This work was supported by National Key R\&D Program of China under no. 2018YFC0807500, National Natural Science Foundation of China under nos. 61832016, 61720106006 and 61672520, as well as CASIA-Tencent Youtu joint research project.






\bibliography{meta}
\bibliographystyle{icml2019}

\end{document}